%% file: iclr2026_conference.tex
\documentclass{article} 
\usepackage{iclr2026_conference,times}

\input{math_commands.tex}

\usepackage{hyperref}
\usepackage{url}
\usepackage[utf8]{inputenc}
\usepackage[T1]{fontenc}
\usepackage{url}
\usepackage{booktabs}
\usepackage{colortbl}
\usepackage{amsfonts}
\usepackage{nicefrac}
\usepackage{microtype}
\usepackage{xcolor}
\usepackage{graphicx}
\usepackage{subfig}
\usepackage{amsmath, amssymb}
\usepackage{wrapfig}
\usepackage{pifont}
\usepackage{caption}
\usepackage{natbib}
\usepackage[most]{tcolorbox}
\usepackage{adjustbox}
\usepackage{makecell} 
\usepackage{multirow}
\usepackage{algorithm}
\usepackage{algpseudocode}
\newcommand{\mycomment}[1]{\hfill\small\texttt{$\triangleright$ #1}}
\usepackage{hyperref}

\usepackage{mathtools}
\usepackage{amsthm}
\usepackage{arydshln} 
\definecolor{colorTab}{rgb}{0.9,0.9,0.98}
\definecolor{color3}{gray}{0.95}
\usepackage{enumitem}
\usepackage{multicol}

\usepackage[capitalize,noabbrev]{cleveref}
\definecolor{css}{rgb}{0.7529, 0, 0}
\definecolor{fss}{rgb}{0, 0.7, 0.3}
\definecolor{pbp}{rgb}{0.2, 0.4, 0.9}

\theoremstyle{plain}
\newtheorem{theorem}{Theorem}[section]

\theoremstyle{definition}

\theoremstyle{remark}

\definecolor{navyblue}{rgb}{0.0, 0.0, 0.5}
\hypersetup{
colorlinks=true,
linkcolor=red,
citecolor=navyblue,
filecolor=navyblue,
urlcolor=navyblue}

\title{Progressive Binarization with Semi-Structured Pruning for LLMs}

\author{
  Xianglong Yan$^{1}$\thanks{Equal contribution.},\enspace 
  Tianao Zhang$^{1}$\footnotemark[1],\enspace
  Zhiteng Li$^{1}$,\enspace
  Haotong Qin$^{2}$,\enspace
  \textbf{Yulun Zhang$^{1}$\thanks{Corresponding author: yulun100@gmail.com}}\enspace \\
  $^{1}$Shanghai Jiao Tong University,\enspace
  $^{2}$ETH Z\"{u}rich
}

\iclrfinalcopy 
\begin{document}

\maketitle
\vspace{-6mm}
\begin{abstract}
\vspace{-4mm}
Large language models (LLMs) have achieved remarkable progress in natural language processing, but their high computational and memory costs hinder deployment on resource-constrained devices. Binarization represents the most extreme form of quantization, yet binarized models still contain redundancy that can be further removed. Pruning provides a natural way to eliminate such redundancy, but naïve combination with binarization often results in severe performance degradation. In this paper, we propose Progressive Binarization with Semi-Structured Pruning (PBS$^2$P), a novel post-training framework that seamlessly integrates binarization and semi-structured pruning. We first propose Stepwise semi-structured Pruning with Binarization Optimization (SPBO), which progressively introduces sparsity while optimizing binarization parameters to jointly reduce pruning and quantization error, yielding more stable and accurate compression. Additionally, we propose a Coarse-to-Fine Search (CFS) that first allocates pruning ratios and then refines element selection, further enhancing overall performance. Extensive experiments across multiple LLM families show that PBS$^2$P consistently outperforms state-of-the-art (SOTA) binary post-training quantization methods in both perplexity and downstream accuracy. The code and models will be available at \url{https://github.com/XIANGLONGYAN/PBS2P}. 
\end{abstract}
\setlength{\abovedisplayskip}{2pt}
\setlength{\belowdisplayskip}{2pt}
\vspace{-5mm}
\section{Introduction}
\vspace{-3mm}
\begin{wrapfigure}{r}{0.5\textwidth}
\vspace{-3.3mm}
 \centering
\includegraphics[trim=3 0 0 0, clip, width=0.5\textwidth]{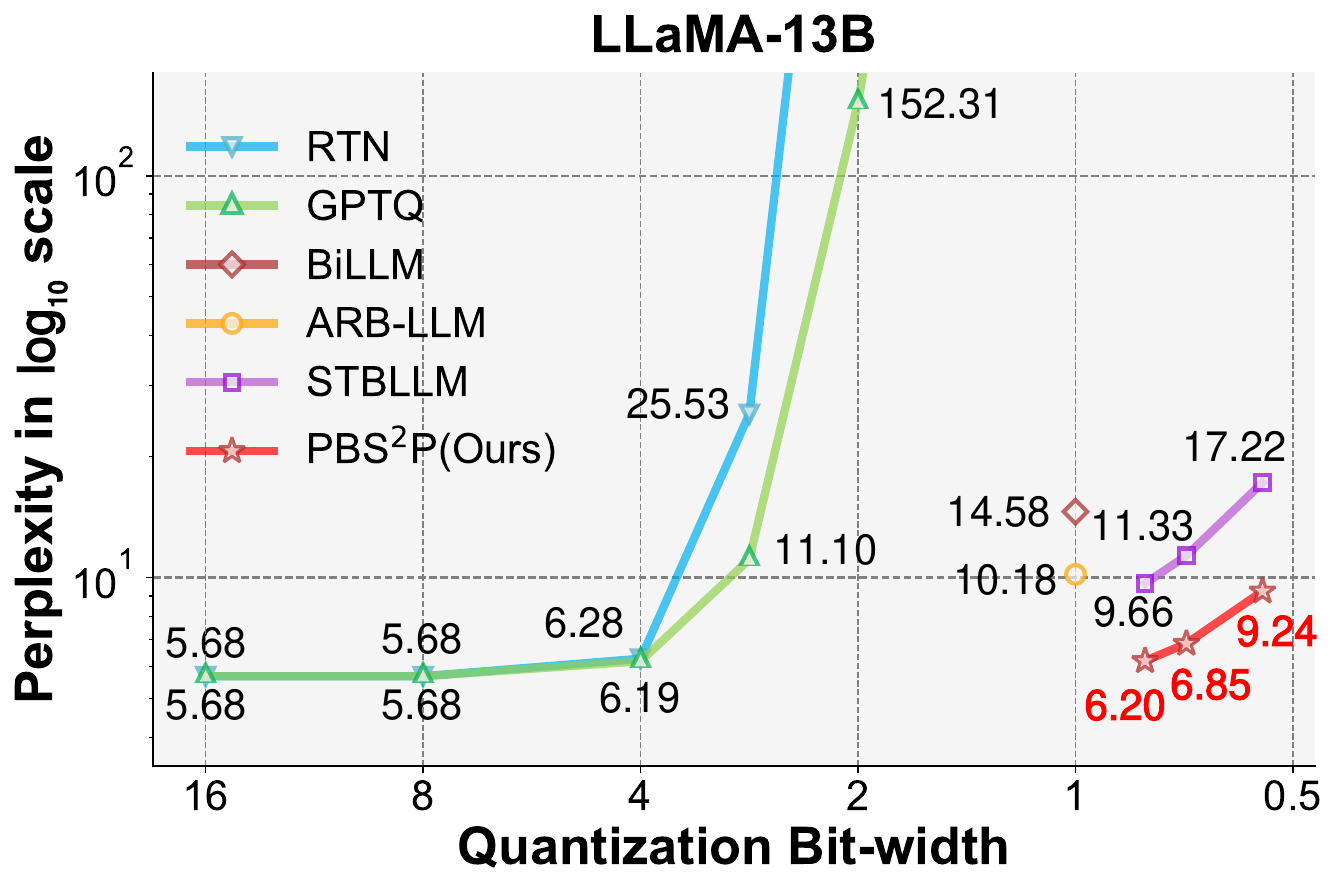}
\vspace{-4.4mm} 
\caption{Perplexity of LLaMA-13B on WikiText2 under different bit-widths across various methods.}
\label{fig:binary_results}
\vspace{-7.5mm}
\end{wrapfigure} 
Transformer-based large language models (LLMs)~\citep{vaswani2017attention} have demonstrated remarkable performance across a wide range of natural language processing (NLP) tasks. This success is largely driven by their massive scale, often with billions of parameters. For instance, the OPT series~\citep{zhang2022opt} includes models with up to 66 billion parameters, while the LLaMA family~\citep{touvron2023llama1} features even larger models, such as LLaMA3-70B~\citep{dubey2024llama3}. Although these models have substantially advanced the state of the art (SOTA) in NLP, their growing computational and memory demands present increasingly significant challenges for real-world deployment.

The compression of LLMs can be broadly categorized into several approaches, including weight quantization~\citep{lin2024awq, frantar2022gptq}, low-rank factorization~\citep{zhang2023loraprune, yuan2023asvd}, network pruning~\citep{sun2023simple, frantar2023sparsegpt}, and knowledge distillation~\citep{zhong2024revisiting, gu2023knowledge}. Among these, binarization represents an extreme form of quantization, reducing model weights to 1 bit and drastically lowering memory consumption. Recent efforts primarily focus on post-training quantization (PTQ)~\citep{huang2024billm, li2024arb, dong2024stbllm}, which eliminates the need for backpropagation, thereby accelerating the binarization process and improving deployment efficiency. For example, BiLLM~\citep{huang2024billm} introduces a residual approximation strategy to enhance the performance of 1-bit LLMs, while ARB-LLM~\citep{li2024arb} adopts an alternating refinement approach to better align binarized weights with their full-precision counterparts. STBLLM~\citep{dong2024stbllm} further advances this line of work by compressing LLMs to sub-1-bit precision. Despite the aforementioned advancements, binarized LLMs continue to exhibit redundancy, indicating opportunities for additional compression.

Pruning~\citep{lecun1989optimal} is a promising technique for further reducing redundancy in binarized models. However, conventional structured pruning~\citep{ma2023llm, ashkboos2024slicegpt, xia2023sheared, an2024fluctuation} often leads to severe performance degradation in LLMs. On the other hand, unstructured pruning~\citep{dong2024pruner} suffers from poor hardware compatibility and inefficient storage. Semi-structured pruning~\citep{frantar2023sparsegpt, sun2023simple, dong2024stbllm} has emerged as an effective compromise, significantly reducing redundancy while maintaining a balance between accuracy and hardware efficiency. Nonetheless, directly combining binarization with semi-structured pruning often leads to noticeable performance degradation, as the joint constraints on weight values and structure increase the difficulty of preserving model accuracy under extreme compression. This problem is further complicated by the need to select pruning elements carefully, since suboptimal choices may remove important weights and amplify quantization errors. Therefore, designing a unified framework that balances compression and accuracy retention remains challenging.

To address these challenges, we propose \textbf{P}rogressive \textbf{B}inarization with \textbf{S}emi-\textbf{S}tructured \textbf{P}runing (PBS$^2$P) for LLMs, which achieves substantial model compression while maintaining strong performance (see Figure~\ref{fig:binary_results}). We first propose \underline{S}tepwise semi-structured \underline{P}runing with \underline{B}inarization \underline{O}ptimization (SPBO), which progressively prunes a subset of elements at each step while jointly optimizing the binarized parameters. This strategy effectively reduces the cumulative error from pruning and binarization. To further enhance pruning efficiency, we develop a \underline{C}oarse-to-\underline{F}ine \underline{S}earch (CFS) algorithm for more accurate pruning element selection. In the coarse stage, pruning ratios are assigned to each layer based on its importance. In the fine stage, a Hessian-based metric is used to identify the most redundant elements, guided by the layer-specific pruning ratios. We further show that the Hessian-based metric minimizes the theoretical error increase under a second-order approximation. This makes it a provably optimal strategy for pruning. Together, SPBO and CFS form the core of PBS$^2$P, enabling provably efficient compression with minimal accuracy degradation.

Extensive experiments demonstrate that PBS$^2$P consistently achieves SOTA performance across multiple LLM families, clearly surpassing existing binary PTQ methods on a wide range of evaluation benchmarks. As shown in Figure~\ref{fig:binary_results}, on the WikiText-2~\citep{merity2016pointer} dataset, PBS$^2$P attains a perplexity of 6.20 on LLaMA-13B~\citep{touvron2023llama1} with an average bit-width of just 0.8 bits, compared to 5.47 for the full-precision model. These results further highlight the effectiveness of PBS$^2$P in significantly narrowing the performance gap between binarized and full-precision models.

Our key contributions can be summarized as follows:
\vspace{-2.5mm}
\begin{itemize}[itemsep=4pt,topsep=6pt,parsep=0pt]
    
    \item We propose a novel framework, PBS$^2$P, which integrates binarization and semi-structured pruning seamlessly for effective LLM compression.
    
    \item We propose \underline{S}tepwise semi-structured \underline{P}runing with \underline{B}inarization \underline{O}ptimization (SPBO), which progressively prunes the model while jointly optimizing binarization parameters, effectively reducing the combined error and preserving performance.

    \item We propose \underline{C}oarse-to-\underline{F}ine \underline{S}earch (CFS) for selecting pruning elements, which enables effective identification of redundant parameters while better preserving model performance.
    
    \item Extensive experiments show that PBS$^2$P consistently outperforms SOTA binary PTQ methods and significantly narrows the gap to full-precision models across diverse benchmarks.

\end{itemize}

\vspace{-6mm}
\section{Related works}
\vspace{-3mm}
\subsection{LLM Quantization}
\vspace{-3mm}
Quantization compresses full-precision parameters into lower-bit representations, reducing both computation and storage demands. Current quantization methods for LLMs are mainly divided into Quantization-Aware Training (QAT) and Post-Training Quantization (PTQ). QAT~\citep{liu2023llm,chen2024db,du2024bitdistiller} integrates quantization during the training phase to enhance low-bit weight representations. However, due to the enormous parameter number, retraining becomes excessively expensive and inefficient for LLMs. PTQ, as it directly applies quantization to the model weights without retraining, making it faster and less resource-demanding. Recent methods, like ZeroQuant~\cite{yao2022zeroquant} and BRECQ~\citep{li2021brecq}, improve quantization accuracy by incorporating custom quantization blocks and group labels. While GPTQ~\citep{frantar2022gptq} and QuIP~\citep{chee2024quip} use second-order error compensation to reduce quantization errors. 
\vspace{-2mm}
\subsection{Network Binarization}
\vspace{-2mm}
Binarization, as the most extreme form of quantization, reduces model parameters to a single bit (±1). Prominent methods, like Binary Weight Network (BWN)~\citep{rastegari2016xnor}, XNOR-Net~\citep{rastegari2016xnor} and Bi-Real Net\citep{liu2018bi}, focus on binarizing the weights, with XNOR-Net~\citep{rastegari2016xnor} and Bi-Real Net\citep{liu2018bi} also binarizing activations. In the context of LLM binarization, BitNet~\citep{wang2023bitnet}, OneBit~\citep{xu2024onebit}, and BinaryMoS~\citep{jo2024mixture} adopt the QAT framework, while BiLLM~\citep{huang2024billm}, ARB-LLM~\citep{li2024arb}, and STBLLM\citep{dong2024stbllm} use PTQ combined with residual approximation. Focusing on the PTQ setting, our method yields significant enhancements over prior SOTA binary PTQ techniques.
\vspace{-6mm}
\subsection{Model Pruning} 
\vspace{-2mm}
Pruning~\citep{lecun1989optimal} is a common technique for compressing neural networks by removing less important parameters, resulting in smaller and more efficient sparse models. In LLMs, pruning methods are typically categorized as structured, unstructured, or semi-structured. Structured pruning~\citep{ma2023llm,ashkboos2024slicegpt,xia2023sheared,an2024fluctuation} removes entire components to enhance efficiency but often causes performance drops, requiring retraining. Unstructured pruning~\citep{dong2024pruner} eliminates individual weights based on importance, preserving performance even at high sparsity, though the irregular patterns are hardware-unfriendly. Semi-structured pruning offers a good trade-off, enforcing regular patterns like $N$:$M$ sparsity for hardware efficiency, as used in SparseGPT~\citep{frantar2023sparsegpt}, Wanda~\citep{sun2023simple}, and STBLLM~\citep{dong2024stbllm}. We adopt $N$:$M$ sparsity to balance performance and hardware efficiency.
\begin{figure}[t]
\centering
\includegraphics[width=\linewidth]{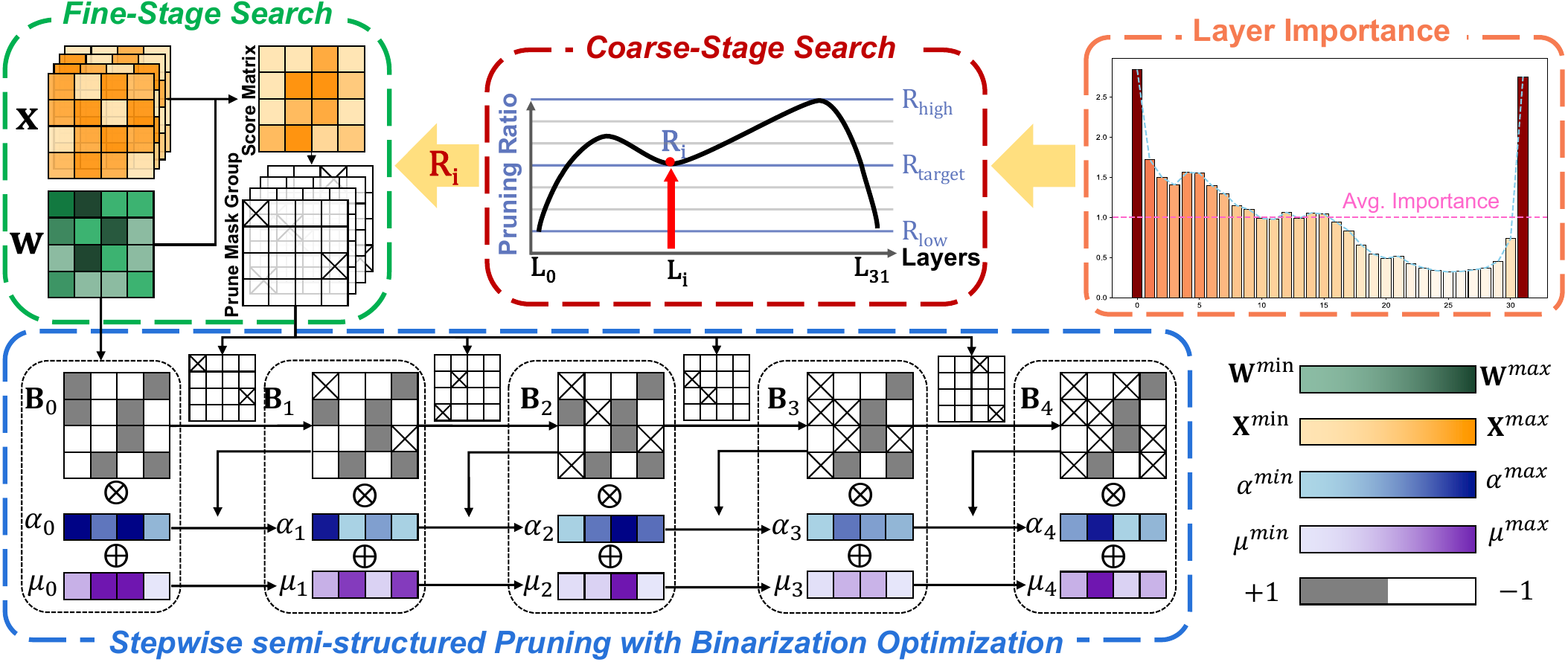}
\vspace{-6mm}
\caption{Overview of our PBS$^2$P framework. 
\textcolor{css}{\textbf{Coarse-Stage Search}}: using the layer importance to assign pruning ratios to each layer. 
\textcolor{fss}{\textbf{Fine-Stage Search}}: searching the elements to be pruned based on the Hessian-based metrics. 
\textcolor{pbp}{\textbf{Stepwise semi-structured Pruning with Binarization Optimization}}: stepwise pruning with optimization of binarized parameters.}
\label{main_pbpllm}
\vspace{-5mm}
\end{figure}
\vspace{-2mm}
\section{Methodology}
\vspace{-2mm}
\label{section3}
\subsection{Preliminary}\label{observation}
\vspace{-2mm}
\textbf{Binarization.$\quad$} We begin by revisiting the standard binarization process~\citep{rastegari2016xnor}. The full-precision matrix $\mathbf{W} \in \mathbb{R}^{n \times m}$ is first row-wise normalized by subtracting the mean of each row, yielding a zero-centered matrix $\widetilde{\mathbf{W}} \in \mathbb{R}^{n \times m}$:
\begin{equation} \label{eq:1}
    \widetilde{\mathbf{W}} = \mathbf{W} - \mu, \quad \text{where} \quad \mu = \frac{1}{m} \sum_{j=1}^m \mathbf{W}_{.j}\, .
\end{equation}
This preprocessing step mitigates distributional bias and promotes a more symmetric weight distribution, which is beneficial for subsequent binarization. The goal of binarization is to minimize the quantization error: $\arg\min_{\alpha,\mathbf{B}} \| \widetilde{\mathbf{W}} - \alpha \mathbf{B} \|_F^2$, where \( \alpha \in \mathbb{R}^n \) is a row-wise scaling vector and \( \mathbf{B} \in \{+1, -1\}^{n \times m} \) is the binary matrix. The closed-form solutions are given by \( \alpha = \frac{1}{m} \sum_{j=1}^m |\widetilde{\mathbf{W}}_{.j}| \), which computes the mean absolute value across each row, and \( \mathbf{B} = \operatorname{sign}(\widetilde{\mathbf{W}}) \), which binarizes the weights via sign function~\citep{huang2024billm}. 

\textbf{N:M Sparsity.$\quad$}
$N$:$M$ sparsity is a semi-structured pruning scheme that balances model compression and hardware efficiency. Unlike unstructured pruning, which removes arbitrary weights and results in irregular sparsity patterns that are hard to accelerate, $N$:$M$ sparsity imposes a structured constraint by retaining exactly $N$ nonzero elements out of every $M$ consecutive weights~\citep{frantar2023sparsegpt,sun2023simple,dong2024stbllm}. A typical configuration, such as 2:4 sparsity, keeps only two weights out of every four, significantly reducing parameter count while preserving computational efficiency. This regular pattern allows for efficient execution on NVIDIA Ampere GPUs~\citep{nvidia2020}, taking advantage of specialized hardware support. In our work, we adopt the $N$:$M$ sparsity scheme as a semi-structured pruning strategy to further eliminate redundancy within the binarized model. This enables stronger compression with hardware-friendly execution.
\vspace{-4mm}
\subsection{Stepwise Semi-Structured Pruning with Binarization Optimization}\label{PBP}
\vspace{-2mm}
To facilitate the joint optimization of binarization and semi-structured pruning, we formulate the total compression error $\mathcal{L}$ as the objective function to be minimized:
\begin{equation}
    \label{eq:layerwise-pruning}   
    \mathcal{L}=||\mathbf{W} \mathbf{X} - (\mathbf{M}_{\text{p}} \odot  \mathbf{\widehat{W}}) \mathbf{X}||_F^2,\quad \text{where}\, \widehat{\mathbf{W}}=\alpha\mathbf{B}+\mu\,.
\end{equation}
where $\mathbf{X}$ denotes the calibration data, $\mathbf{M}_\text{p}$ denotes the semi-structured pruning mask. Optimizing both the pruning mask $\mathbf{M}_{\text{p}}$ and the binarized parameters $\mathbf{\widehat{W}}$ simultaneously is an NP-hard problem~\citep{blumensath2008iterative}. Therefore, we adopt a greedy-style approximation. At each step, we prune a small portion of elements. Then, we update the binarized parameters to fit the current pruning ratio. This step-by-step strategy reduces the optimization difficulty. It distributes optimization across multiple pruning stages. Selection of the semi-structured mask $\mathbf{M}_\text{p}$ is detailed in Section~\ref{C2FS}.

\begin{wrapfigure}{r}{0.5\textwidth}
\vspace{-2.5mm}
 \centering
\includegraphics[trim=3 0 0 0, clip, width=0.5\textwidth]{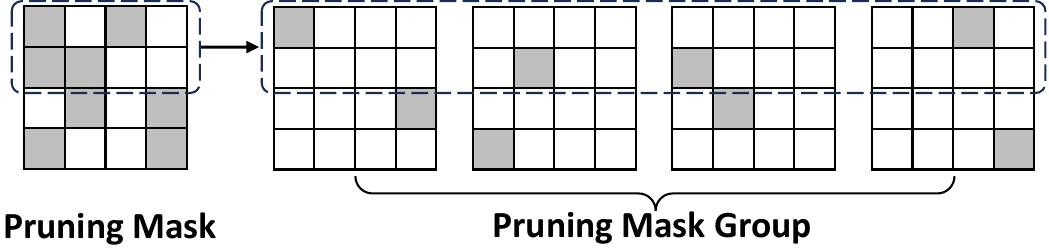}
\caption{Illustration of the $N$:$M$ sparsity pruning mask group, where the original pruning mask $\mathbf{M}_\text{p}$ is divided into $M$-$N$ sub masks.}
\label{fig:prune_mask}
\vspace{-6mm}
\end{wrapfigure} 

\textbf{N: M Sparsity Pruning Mask Group.} 
To support progressive pruning, we decompose the original pruning mask $\mathbf{M}_{\text{p}}$ into a sequence of sub-masks, each corresponding to one intermediate pruning step. Under the $N$:$M$ sparsity constraint, exactly $N$ out of every $M$ elements are retained. Within each $M$-element group, the $M{-}N$ pruned elements are removed sequentially across successive stages. Formally, the full pruning mask can be expressed as the union of $M{-}N$ sub-masks: $\mathbf{M}_{\text{p}} = \bigcup_{i=1}^{M-N} \mathbf{M}_{\text{p}}^i$. This decomposition enables fine-grained progressive pruning and allows for a tighter and more seamless integration with binarization in subsequent stages, as illustrated in Figure~\ref{fig:prune_mask}.

\textbf{Stepwise Pruning with Binarization Optimization (SPBO).$\quad$} 
After obtaining the group of pruning masks, we proceed with the stepwise pruning and binarization optimization process. We perform a total of $M{-}N$ pruning steps to achieve $N$:$M$ sparsity. After each step, we update the binarization parameters $\alpha$ and $\mu$ to better adapt to the current pruning state. At the \(k\)-th pruning step, we obtain the updated binary matrix \(\mathbf{B}_k\) and full-precision matrix \(\mathbf{W}_k\) by applying the \(k\)-th pruning sub-mask \(\mathbf{M}_\text{p}^k\) to the previously pruned matrices \(\mathbf{B}_{k-1}\) and \(\mathbf{W}_{k-1}\):
\begin{equation}
   \mathbf{B}_k = \mathbf{B}_{k-1}\odot\mathbf{M}_\text{p}^k, \quad\mathbf{W}_k = \mathbf{W}_{k-1}\odot\mathbf{M}_\text{p}^k.
\end{equation}

After the $k$-th pruning step, the binarization parameters $\alpha$ and $\mu$ are no longer optimal under the updated sparsity pattern. To minimize the error $\mathcal{L}$ in the current pruning state, these parameters must be re-estimated. Inspired by~\cite{li2024arb}, a straightforward approach is to compute the partial derivatives of $\mathcal{L}$ with respect to $\alpha$ and $\mu$, and solve for the stationary points by setting them to zero. Specifically, we first update $\mu$ by setting $\partial \mathcal{L} / \partial \mu = 0$ during the $k$-th pruning step:
\begin{align}
\partial \mathcal{L} / \partial \mu=0\quad\Rightarrow\quad\mu = \frac{\mathbf{1}^\top \mathbf{S} (\mathbf{W}_k - \alpha \mathbf{B}_k)^\top}{\mathbf{1}^\top \mathbf{S} \mathbf{1}},
\end{align}
where $\mathbf{S} = \sum\nolimits_b \mathbf{X}_b^\top \mathbf{X}_b$ is a precomputed matrix derived from the calibration data $\mathbf{X}$ for notational convenience. Then, we adjust  $\alpha$  by setting $\partial \mathcal{L} / \partial \alpha=0$:
\begin{align}
\partial \mathcal{L} / \partial \alpha=0\quad\Rightarrow\quad\alpha= \frac{\operatorname{diag}(\mathbf{B}_k\mathbf{S}(\mathbf{W}_k-\mu)^\top)}{\operatorname{diag}(\mathbf{B}_k\mathbf{S}\mathbf{B}_k^\top)}.
\end{align}
During the $k$-th pruning round, we can alternately update $\mu$ and $\alpha$ for \(\tau\) steps to further reduce $\mathcal{L}$. We perform total $M$-$N$ steps of pruning until all pruning elements are removed. The detailed derivation of optimizing $\mu$ and $\alpha$ can be found in the supplementary material.

\subsection{Coarse-to-Fine Search}\label{C2FS}
Selecting effective pruning elements is a key challenge in model pruning. To address this, we adopt a two-stage strategy. In the coarse stage, we assign pruning ratios to each layer based on its layer importance. In the fine stage, we determine the specific elements to prune within each layer according to the assigned ratios, allowing for more accurate selection of pruning candidates.

\textbf{Coarse-Stage: Pruning Ratio Allocation.$\quad$}
To estimate the importance of each layer, inspired by~\cite{dumitru2024change}, we use cosine similarity to measure the alignment between its input and output representations. We adopt a straightforward criterion that quantifies how strongly a layer shapes the input–output relationship. Layers that induce larger changes are regarded as more important, whereas those with minimal effect are considered less important. Specifically, for a given weight layer \(\mathcal{W}\), let \(\mathcal{X}_{\mathcal{W}}\) and \(\mathcal{Y}_{\mathcal{W}}\) denote its input and output, respectively. The importance of the layer is then computed as follows:
\begin{equation}
\text{Importance}(\mathcal{W}) = \frac{1}{\text{similarity}(\mathcal{X_{\mathcal{W}},Y_{\mathcal{W}}})}, \quad \text{where}\, \,\,\text{similarity}(\mathcal{X_{\mathcal{W}},Y_{\mathcal{W}}}) = \frac{\mathcal{X}_{\mathcal{W}} \cdot \mathcal{Y}_{\mathcal{W}}}{\|\mathcal{X}_{\mathcal{W}}\|_2 \|\mathcal{Y}_{\mathcal{W}}\|_2}.
\end{equation}
Based on the computed importance, we rank all layers, with \(k_i\) denoting the rank of the \(i\)-th layer. The pruning parameter \(N_i\) for each layer is determined as follows:
\begin{equation}
    N_i = \left\lfloor N_{\text{high}} - \left( N_{\text{high}} - N_{\text{low}} \right) \cdot \frac{k_i-1}{L-1} + \frac{1}{2} \right\rfloor,
\end{equation}
where \( L \) represents the total number of layers in the model. \( N_{\text{high}} \) is the maximum pruning parameter, and \( N_{\text{low}} \) is the minimum pruning parameter. Both are predefined constants. For example, suppose the \(m\)-th layer is the most important with \(k_m = 1\). It will be assigned \(N_\text{high}\), retaining more elements. In contrast, if the \(n\)-th layer is the least important with \(k_n = L\), it will be assigned \(N_\text{low}\), preserving the fewest elements. We define the average pruning parameter \( N_{\text{average}} \) as the average of the maximum and minimum pruning parameters, such that \( N_{\text{average}} = (N_{\text{high}} + N_{\text{low}})/2 \).

\vspace{1mm}
\textbf{Fine-Stage: Selecting Pruning Elements.$\quad$} Once the pruning ratio $N/M$ is assigned in the coarse-stage, we proceed to search for the specific pruning elements within each layer. We revisit the impact of pruning on model error. Inspired by~\cite{hassibi1993optimal}, we derive the relationship between pruning and error increase under a second-order approximation, and present the corresponding formulation in Theorem~\ref{theo}. The full proof is provided in the supplementary material. Since our objective is to minimize the increase in error caused by pruning, we directly adopt the following score to guide element selection: $s_q = \frac{w_q^2}{\left[\mathbf{H}^{-1}\right]_{qq}^2}$. This criterion is theoretically guaranteed to minimize the error increase introduced by pruning under a second-order approximation. Based on the given $N$:$M$ ratio and the corresponding  $s_q$  values, we select the top $N$ elements with the largest  $s_i$ values from every $M$ consecutive elements to retain. The other $M-N$ elements are selected for pruning.

\begin{tcolorbox}[colback=yellow!10, colframe=black, boxrule=0.2 mm, drop shadow]
\begin{theorem}\label{theo}
After pruning an element $w_q$, the pruning-induced error increase $\delta\mathcal{L}$ can be approximated by the following expression:
\begin{equation}
    \delta\mathcal{L}=\frac{1}{2}\frac{w_q^2}{[\mathbf{H}^{-1}]_{qq}},
\end{equation}
where $[\mathbf{H}^{-1}]_{qq}$ denotes the $q$-th diagonal entry of the inverse Hessian matrix.
\end{theorem}
\end{tcolorbox}

\begin{algorithm*}[t]
\caption{Pseudocode of PBS$^2$P}
\begin{algorithmic}[1]
\State \textbf{Inputs:} Model $\mathcal{M}$, Calibration Data $\mathcal{X}$, Target Pruning Ratio $\mathcal{TR}$, Optimization Steps $\tau$
\State \textbf{Output:} Binarized and Pruned Model $\mathcal{M}'$
\Procedure{PBS$^2$P}{$\mathcal{M}, \mathcal{X}, \mathcal{TR},\tau$}

\State $R \gets$ \textproc{Coarse\_Stage\_Search}$(\mathcal{M}, \mathcal{X}, \mathcal{TR})$ \mycomment{Allocate layer-wise pruning ratios}

\For{each layer $l$ in model $\mathcal{M}$}
    \State $\mathcal{W}_l \gets$ \textproc{Extract\_Weights}$(\mathcal{M}, l)$ \mycomment{Obtain weights of layer $l$}
    \State $r_l \gets R[l]$ \mycomment{Get pruning ratio for layer $l$}
    \State $\mathbf{M}_\text{p} \gets$ \textproc{Fine\_Stage\_Search}$(\mathcal{W}_l, \mathcal{X}, r_l)$ \mycomment{Generate pruning mask}
    \State $\mathcal{W}_l' \gets$ \textproc{SPBO}$(\mathcal{W}_l, \mathcal{X}, \mathbf{M}_\text{p}, \tau)$ \mycomment{Stepwise pruning and binarization optimization}
    \State $\mathcal{M}' \gets$ \textproc{Update\_Layer}$(\mathcal{M}, l, \mathcal{W}_l')$ \mycomment{Write updated weights back to model}
\EndFor

\State \Return{$\mathcal{M}'$}
\EndProcedure
\end{algorithmic}
\label{alg:pbsp}
\end{algorithm*}
\vspace{-2mm}
\subsection{PBS$^2$P Pipeline}\label{Pipeline}
\vspace{-2mm}
\textbf{PBS$^2$P Workflow.$\quad$}
As illustrated in Figure~\ref{main_pbpllm}, the proposed PBS$^2$P framework is composed of three key components: Coarse-stage Search, Fine-stage Search, and Stepwise Semi-structured Pruning with Binarization Optimization. The workflow of PBS$^2$P is outlined in Algorithm~\ref{alg:pbsp}. Given a full-precision model $\mathcal{M}$, calibration data $\mathcal{X}$, and a target average pruning ratio $\mathcal{TR}$, the algorithm first performs a coarse-stage search to determine layer-wise pruning ratios based on importance. With the ratios determined, we proceed layer by layer to apply pruning and binarization. For each layer $l$, we extract its weight matrix $\mathcal{W}_l$ and conduct a fine-stage search to generate a semi-structured pruning mask $\mathbf{M}_\text{p}$. The SPBO procedure is then invoked to progressively prune the weights while optimizing the binarization parameters. This joint optimization reduces the total error introduced by pruning and binarization. The updated weights $\mathcal{W}_l'$ are written back to the model, and the process repeats for all layers. As pruning and binarization are applied layer-wise, our method has low memory overhead.

\textbf{Average Bits.$\quad$}
Following BiLLM~\citep{huang2024billm} and STBLLM~\citep{dong2024stbllm}, we similarly divide weights into salient and non-salient groups, consistent with their design. For salient weights, we adopt a residual binarization strategy, allocating two bits to better preserve their expressiveness, as done in both prior works. For non-salient weights, we apply group-wise quantization. In line with STBLLM~\citep{dong2024stbllm}, we further divide the non-salient weights into three groups using two split points, maintaining the same quantization scheme. The weight parameters and additional hardware overhead are as follows:
\begin{equation}
    \left\{
         \begin{array}{lr}
         N_{\text{param}} = \left[2 \times r_{\text{salient}} +(1 - r_{\text{salient}}) \right] \times \frac{N}{M}, \\
         N_{\text{storing}} =  2 + \dfrac{1}{b_{\text{size}}}, \\
         \end{array}
    \right.
\end{equation}
where \( r_{\text{salient}} \) represents the proportion of salient weights, $\frac{N}{M}$ denotes the predefined average pruning ratio for the entire model, and \( b_{\text{size}} \) indicates the block size used in OBC~\citep{frantar2022optimal} compensation, with 2 bits reserved to mark the division between salient and non-salient weights. Our parameter settings and $\frac{N}{M}$ configuration are identical to our main comparison method STBLLM~\citep{dong2024stbllm}. Specifically, we adopt the same settings of 4:8, 5:8, and 6:8 sparsity as used in STBLLM~\citep{dong2024stbllm}. These correspond to average bit-widths of 0.55, 0.70, and 0.80, respectively. This alignment ensures fair, direct comparison under consistent compression budgets.

\vspace{-2mm}
\section{Experiments}
\label{section4}
\vspace{-2mm}

\subsection{Settings}
\vspace{-2mm}
All experiments are performed using PyTorch~\citep{paszke2019pytorch} and Huggingface~\citep{paszke1912imperative} on a single NVIDIA A800-80GB GPU. Following the work of \citep{frantar2022gptq}, \citep{huang2024billm}, and \citep{li2024arb}, we use 128 samples from the C4~\citep{raffel2020exploring} dataset for calibration. Since PBS$^2$P is an efficient PTQ framework, it eliminates fine-tuning, enabling completion through a single process combining binarization and pruning.

\textbf{Models and Datasets.$\quad$}
We conduct extensive experiments on the LLaMA~\citep{touvron2023llama1}, LLaMA-2~\citep{touvron2023llama2}, and LLaMA-3~\citep{dubey2024llama3} families and the OPT family~\citep{zhang2022opt}. To evaluate the effectiveness of PBS$^2$P, we measure the perplexity of LLM outputs on WikiText2~\citep{merity2016pointer}, PTB~\citep{marcus1994penn}, and C4~\citep{raffel2020exploring}. Moreover, we also evaluate the accuracy on seven zero-shot QA datasets: ARC-c~\citep{clark2018think}, ARC-e~\citep{clark2018think}, BoolQ~\citep{clark2019boolq}, Hellaswag~\citep{zellers2019hellaswag}, OBQA~\citep{mihaylov2018can}, RTE~\citep{chakrabarty2021figurative}, and Winogrande~\citep{sakaguchi2019adversarial}. This benchmark setup validates our method on both language modeling and reasoning tasks.

\textbf{Baselines.$\quad$}
We mainly compare our PBS$^2$P with STBLLM~\citep{dong2024stbllm}, a structural binary PTQ framework designed for compressing LLMs to precisions lower than 1-bit. We compare the results of PBS$^2$P with STBLLM under the same $N$:$M$ sparsity settings (\textit{e.g.}, 4:8, 5:8, 6:8). Previous low-bit methods like ARB-LLM~\citep{li2024arb}, BiLLM~\citep{huang2024billm}, GTPQ~\citep{frantar2022gptq}, and vanilla RTN are also selected as baselines for comparison.

\begin{table*}[!t]
\renewcommand{\arraystretch}{1.0}
\footnotesize
\centering
\caption{Perplexity comparison of RTN, GPTQ~\citep{frantar2022gptq}, BiLLM~\citep{huang2024billm}, ARB-LLM~\citep{li2024arb}, STBLLM~\citep{dong2024stbllm}, and PBS$^2$P on the LLaMA families. The evaluation results demonstrate the perplexity performance on the Wikitext2~\citep{merity2016pointer} dataset across various model sizes.}
\label{tab:llama_only}
\vspace{-2mm}
\begin{adjustbox}{width=\textwidth}
\begin{tabular}{|l|c|c|r|r|r|r|r|r|r|}
    \hline
    \rowcolor{color3}
    \multicolumn{3}{|c|}{\textbf{Settings}} & \multicolumn{4}{c|}{\textbf{LLaMA-1}} & \multicolumn{2}{c|}{\textbf{LLaMA-2}} & \multicolumn{1}{c|}{\textbf{LLaMA-3}} \\
    \hline
    \rowcolor{color3}
    Method & \#Block & W-Bits & 7B & 13B & 30B & 65B & 7B & 13B & 8B \\
    \hline
    FP16 & - & 16 & 5.68 & 5.09 & 4.10 & 3.53 & 5.47 & 4.88 & 6.14 \\
    \hline
    RTN & - & 3 & 25.54 & 11.40 & 14.89 & 10.59 & 542.80 & 10.68 & 2194.98 \\
    \hline
    GPTQ & 128 & 3 & 8.63 & 5.67 & 4.87 & 4.17 & 6.44 & 5.46 & 18.68 \\
    \hline
    RTN & - & 2 & 106767.34 & 57409.93 & 26704.36 & 19832.87 & 17788.94 & 51145.61 & 1335816.13 \\
    \hline
    GPTQ & 128 & 2 & 129.20 & 20.46 & 15.29 & 8.66 & 52.22 & 23.63 & 1480.43 \\
    \hline
    RTN & - & 1 & 168388.00 & 1412020.25 & 14681.76 & 65253.24 & 157058.34 & 47902.32 & 1353698.38 \\
    \hline
    GPTQ & 128 & 1 & 164471.78 & 131505.41 & 10339.15 & 20986.16 & 59758.69 & 22926.54 & 1121260.50 \\
    \hline
    BiLLM & 128 & 1.11 & 49.79 & 14.58 & 9.90 & 8.37 & 32.31 & 21.35 & 55.80 \\
    \hline
    ARB-LLM & 128 & 1.11 & 14.03 & 10.18 & 7.75 & 6.56 & 16.44 & 11.85 & 27.42 \\
    \hline
    STBLLM & 128 & 0.80 & 15.03 & 9.66 & 7.56 & 6.43 & 13.06 & 11.67 & 33.44 \\
    \hline
    STBLLM & 128 & 0.70 & 19.48 & 11.33 & 9.19 & 7.91 & 18.74 & 13.26 & 49.12 \\
    \hline
    STBLLM & 128 & 0.55 & 31.72 & 17.22 & 13.43 & 11.07 & 27.93 & 20.57 & 253.76 \\
    \hline
    \rowcolor{colorTab}
    PBS$^2$P & 128 & 0.80 & 7.36 & 6.20 & 5.21 & 4.60 & 7.17 & 6.27 & 10.75 \\
    \hline
    \rowcolor{colorTab}
    PBS$^2$P & 128 & 0.70 & 8.09 & 6.85 & 5.78 & 5.09 & 8.00 & 6.89 & 12.29 \\
    \hline
    \rowcolor{colorTab}
    PBS$^2$P & 128 & 0.55 & 10.78 & 9.24 & 7.19 & 6.39 & 10.64 & 8.68 & 17.45 \\
    \hline
\end{tabular}
\end{adjustbox}
\vspace{-4mm}
\end{table*}

\subsection{Main Results}
We perform a comprehensive comparison of different LLM families (like LLaMA~\citep{touvron2023llama1,dubey2024llama3,touvron2023llama2} and OPT~\citep{zhang2022opt}) with various model sizes. To keep fairness, we follow STBLLM~\citep{dong2024stbllm} to report the average bit-width of all methods, where our methods have the same bit-width as STBLLM. 

\textbf{Results on LLaMA Family.$\quad$}
As shown in Table~\ref{tab:llama_only}, the models using RTN and GPTQ methods find it hard to maintain model performance at 1-bit precision. BiLLM achieves a satisfactory perplexity of 1.11 bits but performs worse than ARB-LLM at the same bit-width. At sub-1-bit precision, PBS$^2$P surpasses STBLLM and significantly reduces perplexity at the same bit-width across model sizes from 7B to 65B. For instance, PBS$^2$P achieves a substantial improvement over STB-LLM on LLaMA-1-7B, with perplexity dropping from 31.72 to 10.78, a reduction of approximately 66.0\%, in the extreme case of 4:8 structured binarization, where half of the parameters are pruned. 

\begin{wraptable}{r}{0.5\textwidth}
\vspace{-4.5mm}
\captionsetup{width=0.5\textwidth}
\caption{Perplexity on the OPT family using Wiki2.}
\label{tab:opt_only}
\centering
\vspace{-1mm}
\begin{adjustbox}{width=0.5\textwidth}
\begin{tabular}{|l|c|c|r|r|r|}
\hline
\rowcolor{color3}
\textbf{Method} & \textbf{Block} & \textbf{W-Bit} & \textbf{1.3B} & \textbf{2.7B} & \textbf{30B} \\
\hline
FP16 & - & 16 & 14.62 & 12.47 & 9.56 \\
\hline
RTN & - & 2 & 11272.65 & 9505.76 & 1165864.25 \\
\hline
GPTQ & 128 & 2 & 121.64 & 59.53 & 13.04 \\
\hline
RTN & - & 1 & 17165.72 & 36516.69 & 6485.99 \\
\hline
GPTQ & 128 & 1 & 8719.58 & 11700.13 & 14083.15 \\
\hline
BiLLM & 128 & 1.11 & 69.05 & 48.61 & 13.86 \\
\hline
ARB-LLM & 128 & 1.11 & 26.63 & 19.84 & 11.12 \\
\hline
STBLLM & 128 & 0.80 & 29.84 & 17.02 & 12.80 \\
\hline
STBLLM & 128 & 0.70 & 33.01 & 20.82 & 14.38 \\
\hline
STBLLM & 128 & 0.55 & 45.11 & 30.34 & 18.80 \\
\hline
\rowcolor{colorTab}
PBS$^2$P & 128 & 0.80 & 23.72 & 18.32 & 10.50 \\
\hline
\rowcolor{colorTab}
PBS$^2$P & 128 & 0.70 & 27.10 & 20.17 & 10.82 \\
\hline
\rowcolor{colorTab}
PBS$^2$P & 128 & 0.55 & 44.53 & 27.42 & 11.87 \\
\hline
\end{tabular}
\end{adjustbox}
\vspace{-4.5mm}
\end{wraptable}

\begin{figure}[!t]
\centerline{\includegraphics[width=\textwidth]{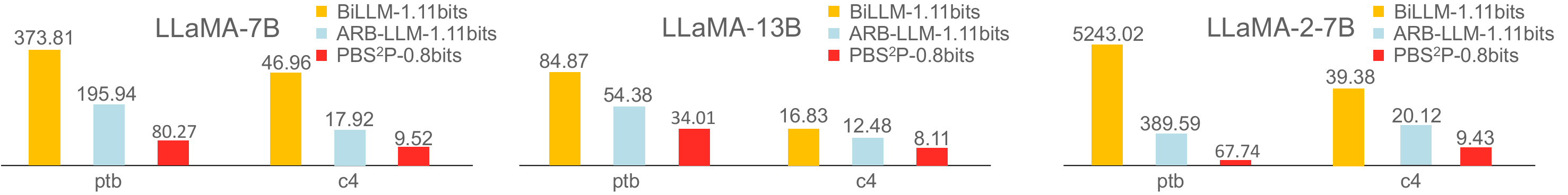}}
\vspace{-2mm}
\caption{BiLLM~\citep{huang2024billm}, ARB-LLM~\citep{li2024arb}, and PBS$^2$P performed on the PTB~\citep{marcus1994penn}and c4~\citep{raffel2020exploring} datasets, mainly on LLaMA-7B, LLaMA-13B, and LLaMA2-7B, and we found that PBS$^2$P at 0.8 bits outperforms other methods.}
\label{ptb_c4}
\vspace{-6mm}
\end{figure}

Furthermore, PBS$^2$P, with a precision of 0.8 bits, outperforms both RTN at 3 bits, GPTQ at 2 bits, BiLLM and ARB-LLM at 1.11 bits, and STBLLM at 0.8 bits in terms of perplexity across all model sizes. Those comparisons show that our PBS$^2$P achieves a better trade-off between bit precision and performance. It is worth noting that PBS$^2$P outperforms GPTQ at 3 bits on LLaMA 1-7B and LLaMA 3-8B. We extend perplexity evaluation to the PTB and C4 datasets. Figure~\ref{ptb_c4} shows the performance of the LLaMA-7B, LLaMA-13B, and LLaMA-2-7B models. PBS$^2$P continues to achieve a leading edge in performance while consistently operating at a relatively lower bit-width compared to other methods across diverse evaluation settings.

\textbf{Results on OPT Family.$\quad$} We extend our experiments to the OPT family (1.3B to 30B) under sub-1-bit PTQ settings, similar to the setup for LLaMA family. As shown in Table~\ref{tab:opt_only}, PBS$^2$P continues to outperform STBLLM across most of the models and $N$:$M$ structured binarization configurations. More results are provided in the supplementary material.

\vspace{-3mm}
\subsection{Zero-Shot Results}
\vspace{-3mm}
 To provide a more thorough evaluation of binary LLMs, we extend our experiments to 7 zero-shot datasets and test on models from the LLaMA family: LLaMA-1-13B, LLaMA-2-13B, and LLaMA-1-30B. Each model is evaluated across various compression methods, including full-precision, STBLLM (6:8), STBLLM (4:8), PBS$^2$P (6:8), and PBS$^2$P (4:8). As shown in Table~\ref{tab:zero_shot_acc}, models compressed with PBS$^2$P significantly outperform those compressed with STBLLM in terms of average accuracy. Such comparisons demonstrate that PBS$^2$P provides a more effective solution for compressing LLMs to less than 1-bit, offering improved trade-offs between accuracy and efficiency. This highlights its strong potential for enabling high-performance, low-bit inference in real-world applications.

\begin{table*}[!t]
\centering
\caption{
Accuracies (\%) for 7 zero-shot tasks from semi-structured binarized LLaMA-1-13B, LLaMA-2-13B, and LLaMA-1-30B with STBLLM and PBS$^2$P.}
\vspace{-2mm}
\begin{adjustbox}{width=1.0\textwidth}
\setlength{\tabcolsep}{5.5pt}
\begin{tabular}{|c|c|c|c|c|c|c|c|c|c|c|}
\hline
\rowcolor{color3}
\textbf{Models} & \textbf{Method} &\textbf{W-Bits}& \textbf{Winogrande} & \textbf{OBQA}  & \textbf{Hellaswag} & \textbf{Boolq} & \textbf{ARC-e}  & \textbf{ARC-c}  & \textbf{RTE}   & \textbf{Average} \\
\hline

\multirow{5}{*}{\textbf{LLaMA-1-13B}} 
& FP16 &16 & 72.69  & 33.20  & 59.91  & 77.89  & 77.40  & 46.42  & 70.40  & 63.80  \\
& STBLLM&0.80 & 65.98  & 36.20  & 63.67  & 65.38  & 68.86  & 34.04  & 56.68  & 55.83  \\
& STBLLM&0.55 & 63.06  & 34.80  & 52.65  & 62.48  & 56.90  & 28.33  & 52.71  & 50.13  \\
& \cellcolor{colorTab}PBS$^2$P &\cellcolor{colorTab}0.80& \cellcolor{colorTab}72.77 & \cellcolor{colorTab}31.00 & \cellcolor{colorTab}54.80 & \cellcolor{colorTab}74.71 & \cellcolor{colorTab}74.37 & \cellcolor{colorTab}42.32 & \cellcolor{colorTab}68.23 & \cellcolor{colorTab}59.74  \\
& \cellcolor{colorTab}PBS$^2$P&\cellcolor{colorTab}0.55 &\cellcolor{colorTab}69.30 & \cellcolor{colorTab}26.80 & \cellcolor{colorTab}46.83 & \cellcolor{colorTab}71.56 & \cellcolor{colorTab}65.70 & \cellcolor{colorTab}32.68 & \cellcolor{colorTab}55.96 & \cellcolor{colorTab}52.69  \\
\hline

\multirow{5}{*}{\textbf{LLaMA-2-13B}} 
& FP16&16 & 72.22  & 35.20  & 60.03  & 80.55  & 79.42  & 48.38  & 65.34  & 65.00  \\
& STBLLM&0.80 & 63.93  & 37.00  & 57.76  & 71.53  & 60.56  & 31.99  & 54.15  & 53.85  \\
& STBLLM&0.55 & 55.88  & 29.40  & 44.03  & 64.31  & 48.86  & 26.54  & 52.71  & 45.96  \\
& \cellcolor{colorTab}PBS$^2$P &\cellcolor{colorTab}0.80 & \cellcolor{colorTab}72.45 & \cellcolor{colorTab}31.00 & \cellcolor{colorTab}54.43 & \cellcolor{colorTab}80.61 & \cellcolor{colorTab}74.28 & \cellcolor{colorTab}42.75 & \cellcolor{colorTab}59.21 & \cellcolor{colorTab}59.24 \\
& \cellcolor{colorTab}PBS$^2$P &\cellcolor{colorTab}0.55 & \cellcolor{colorTab}69.85 & \cellcolor{colorTab}27.00 & \cellcolor{colorTab}47.75 & \cellcolor{colorTab}75.50 & \cellcolor{colorTab}69.19 & \cellcolor{colorTab}35.58 & \cellcolor{colorTab}62.82 & \cellcolor{colorTab}55.38 \\
\hline

\multirow{5}{*}{\textbf{LLaMA-1-30B}} 
& FP16&16 & 75.77  & 36.00  & 63.37  & 82.69  & 80.30  & 52.90  & 67.15  & 67.40  \\
& STBLLM &0.80 & 71.59 & 41.00 & 69.85 & 77.37 & 71.55 & 41.30 & 48.01 & 60.10 \\
& STBLLM &0.55 & 64.01 & 34.60 & 56.46 & 63.06 & 60.86 & 31.48 & 51.99 & 51.78 \\
& \cellcolor{colorTab}PBS$^2$P &\cellcolor{colorTab}0.80 & \cellcolor{colorTab}75.93 & \cellcolor{colorTab}35.00 & \cellcolor{colorTab}59.45 & \cellcolor{colorTab}82.14 & \cellcolor{colorTab}79.29 & \cellcolor{colorTab}47.95 & \cellcolor{colorTab}63.18 & \cellcolor{colorTab}63.28  \\
& \cellcolor{colorTab}PBS$^2$P &\cellcolor{colorTab}0.55 & \cellcolor{colorTab}72.14 & \cellcolor{colorTab}31.20 & \cellcolor{colorTab}53.00 & \cellcolor{colorTab}79.76 & \cellcolor{colorTab}73.99 & \cellcolor{colorTab}41.13 & \cellcolor{colorTab}69.31 & \cellcolor{colorTab}60.08 \\
\hline
\end{tabular}
\end{adjustbox}
\label{tab:zero_shot_acc}
\end{table*}

\vspace{-5mm}
\subsection{Ablation Study}
\vspace{-2mm}
\textbf{Ablation for SPBO Strategy.$\quad$}
To validate the effectiveness of our SPBO strategy, we provide the performance of PBS$^2$P with and without its application. As shown in Table~\ref{tab:Progressive_Strategy}, the performance of SPBO surpasses that of the vanilla pruning followed by binarization approach. In the vanilla method, pruning is done in a single step, where all elements are pruned simultaneously, and binarization is applied to the remaining elements afterward. This approach makes it difficult to jointly minimize the combined pruning and binarization errors. In contrast, our SPBO strategy progressively couples these two processes, resulting in a clear performance improvement.

\textbf{Ablation for Metric in Fine-Stage Search.$\quad$}
The Table~\ref{tab:Fine-Stage_Metric} presents the performance of different pruning metrics in the fine-stage search. We compare several metrics, including random selection, magnitude-based selection, Wanda, SI, and our Hessian-based metric. Our metric significantly outperforms both random selection and magnitude-based pruning, achieving superior performance compared to Wanda and SI, thereby highlighting the effectiveness of our approach. 

\textbf{Ablation for Metric in Coarse-Stage Search.$\quad$}
Table~\ref{tab:Coarse-Stage_Search} shows the performance of the coarse-stage search under three conditions: without the coarse-stage search, using the relative importance (RI) metric from STBLLM, and applying our layer importance (LI) score. The results reveal that using the RI metric causes a performance drop compared to the baseline without coarse-stage search. In contrast, our proposed layer importance (LI) score achieves substantial improvements, validating both the necessity of the coarse-stage search and the superiority of our metric.

\vspace{1mm}
\textbf{Ablation for Pruning Type.$\quad$}
In Table~\ref{tab:Prune_Type}, we present the impact of different pruning types on the results, comparing structured pruning, unstructured pruning, and semi-structured pruning under the same pruning ratio (50\%). We adopt column pruning for the structured case, elementwise magnitude-based pruning for the unstructured case, and 4:8 sparsity for the semi-structured case. Semi-structured pruning strikes an optimal balance: it maintains hardware efficiency while achieving performance close to unstructured pruning and significantly better than structured pruning.

\textbf{Ablation for Group Size.$\quad$}
Table~\ref{tab:group_size} presents the results of our ablation study on the group size configuration. It indicates that a smaller group size, meaning finer-grained grouping, leads to better performance. However, this also comes with increased computational and storage costs. To strike a balance between performance and resource efficiency, we select a group size of 128.

\textbf{Ablation for Calibration Data Size.$\quad$}
Similar to other PTQ-based methods, our PBS$^2$P framework requires only a small calibration set. Table~\ref{tab:Calib_Data_Size} shows the effect of varying calibration size. With the C4 dataset, reducing the sample size to 32 or 64 degrades performance, while increasing to 256 yields minimal additional benefit over 128. PBS$^2$P remains robust with only 128 calibration samples.

\begin{table*}[t]
\vspace{-3mm}
\caption{Ablation study on LLaMA-7B, where all PBS$^2$P is applied an $N$:$M$ sparsity of 4:8. Results are measured by perplexity on the Wikitext2 and C4 datasets. Our results are highlighted in \textbf{bold}.}
\vspace{-2mm}
\subfloat[\small Ablation for SPBO Strategy. \label{tab:Progressive_Strategy}]{\hspace{-1mm}\vspace{-2mm}
\scalebox{0.85}
{\begin{tabular}{c c c}
\toprule
\rowcolor{color3}
\textbf{SPBO Strategy} & \textbf{Wikitext2$\downarrow$} & \textbf{C4$\downarrow$} \\
\midrule
\ding{55} & 14.43 & 16.76 \\
\ding{51} & \textbf{10.78} & \textbf{13.06} \\
\bottomrule
\end{tabular}}
}\hspace{3.2mm}\vspace{-2mm}
\subfloat[\small Ablation for Metric in Fine-Stage Search\label{tab:Fine-Stage_Metric}]
{\scalebox{0.85}
{\begin{tabular}{c c c c c c}
\toprule
\rowcolor{color3}
\textbf{Metric} & \textbf{Random} & \textbf{Magnitude} & \textbf{Wanda} & \textbf{SI} & \textbf{Ours}\\
\midrule
\textbf{Wikitext2$\downarrow$} & 7,779.28 & 38.90 & 10.89 & 196.61 & \textbf{10.78} \\
\textbf{C4$\downarrow$} & 6,797.09 & 24.47 & 13.16 & 148.85 & \textbf{13.06} \\
\bottomrule
\end{tabular}}
}

\subfloat[\small Ablation for Metric in Coarse-Stage Search\label{tab:Coarse-Stage_Search}]{\hspace{-1mm}\vspace{-2mm}
\scalebox{0.77}
{\begin{tabular}{c c c c}
\toprule
\rowcolor{color3}
\textbf{Coarse-Stage Search} & \textbf{Metric} & \textbf{Wikitext2$\downarrow$} & \textbf{C4$\downarrow$} \\
\midrule
\ding{55} & — & 11.55 & 13.94 \\
\ding{51}  & RI & 14.43 & 16.76 \\
\ding{51} & LI & \textbf{10.78} & \textbf{13.06}\\
\bottomrule
\end{tabular}}
}\hspace{3mm}\vspace{-2mm}
\subfloat[\small Ablation for Pruning Type\label{tab:Prune_Type}]
{\scalebox{0.77}
{\hspace{-0.8mm}\begin{tabular}{c c c c}
\toprule
\rowcolor{color3}
\textbf{Prune Type} & \textbf{Hardware-friendly} & \textbf{Wikitext2$\downarrow$} & \textbf{C4$\downarrow$} \\
\midrule
Structured & \ding{51} & 621.16 & 400.50 \\
Unstructured & \ding{55} & 8.54 & 10.95 \\
Semi-Structured & \ding{51} & \textbf{10.78} & \textbf{13.06} \\
\bottomrule
\end{tabular}}
}

\subfloat[\small Ablation for Group Size \label{tab:group_size}]{\hspace{-1mm}\vspace{-5mm}
\scalebox{0.82}
{\hspace{-0.8mm}
\begin{tabular}{c c c c c c c}
\toprule
\rowcolor{color3}
\textbf{Group Size} &\textbf{32} &\textbf{64} & \textbf{128} & \textbf{256} &\textbf{512}\\
\midrule
\textbf{Wikitext2}$\downarrow$& 9.74 & 10.24  & \textbf{10.78} & 11.20 & 11.91  \\
\textbf{C4}$\downarrow$ & 11.75 & 12.50  & \textbf{13.06} & 13.68 & 14.51  \\
\bottomrule
\end{tabular}
}
}\hspace{-3mm}\vspace{-0mm}
\subfloat[\small Ablation for Calibration Data Size\label{tab:Calib_Data_Size}]
{\scalebox{0.82}
{\hspace{4mm}\begin{tabular}{c c c c c}
\toprule
\rowcolor{color3}
\textbf{Calibration Data Size} &\textbf{32}&\textbf{64} & \textbf{128} & \textbf{256} \\
\midrule
\textbf{Wikitext2}$\downarrow$&11.62 & 11.20  & \textbf{10.78} & 10.77\\
\textbf{C4}$\downarrow$ &13.58& 13.44 & \textbf{13.06} & 13.12 \\
\bottomrule
\end{tabular}}
}
\hspace{3mm}
\label{tab:ablations}
\vspace{-9.5mm}
\end{table*}

\vspace{-4mm}
\subsection{Time Analysis}
\vspace{-3mm}
As a binary PTQ framework, PBS$^2$P eliminates the need for fine-tuning and retraining. It introduces coarse-to-fine search and stepwise pruning with binarization optimization, which moderately increases the overall computation time. As shown in Table~\ref{tab:Time_Comparison}, BiLLM~\citep{huang2024billm} completes quantization on LLaMA-7B in 45 minutes, while ARB-LLM~\citep{li2024arb} takes 76 minutes. 
\begin{wraptable}{r}{0.4\textwidth}
\vspace{-3mm}
\captionsetup{width=0.4\textwidth}
\caption{Time comparison of different PTQ methods on LLaMA-7B.}
\label{tab:Time_Comparison}
\centering
\vspace{-2mm}
\scalebox{0.8}{
\hspace{-2mm}
\begin{tabular}{l@{\hskip 6pt}c@{\hskip 6pt}c@{\hskip 6pt}c}
\toprule
\rowcolor{color3}
\textbf{Method} & \textbf{Coarse stage} & \textbf{SPBO} & \textbf{Total} \\
\midrule
BiLLM     & --     & --       & 45 min \\
\midrule
ARB-LLM   & --     & --       & 76 min \\
\midrule
PBS$^2$P  & 2 min  & 109 min  & 111 min \\
\bottomrule
\end{tabular}
}
\vspace{-3mm}
\end{wraptable}
PPBS$^2$P finishes in 111 minutes, with only 2 minutes spent on the coarse-stage search and 109 minutes on the fine-stage search with SPBO. This additional overhead is mainly due to iterative parameter updates in the fine stage. This overhead is acceptable since all computations are carried out offline, imposing no additional latency during inference. With this design, PBS$^2$P achieves superior accuracy and highly efficient, practical deployment.

\begin{wrapfigure}{r}{0.5\textwidth}
\vspace{-4mm}
 \centering
\includegraphics[trim=3 0 0 0, clip, width=0.5\textwidth]{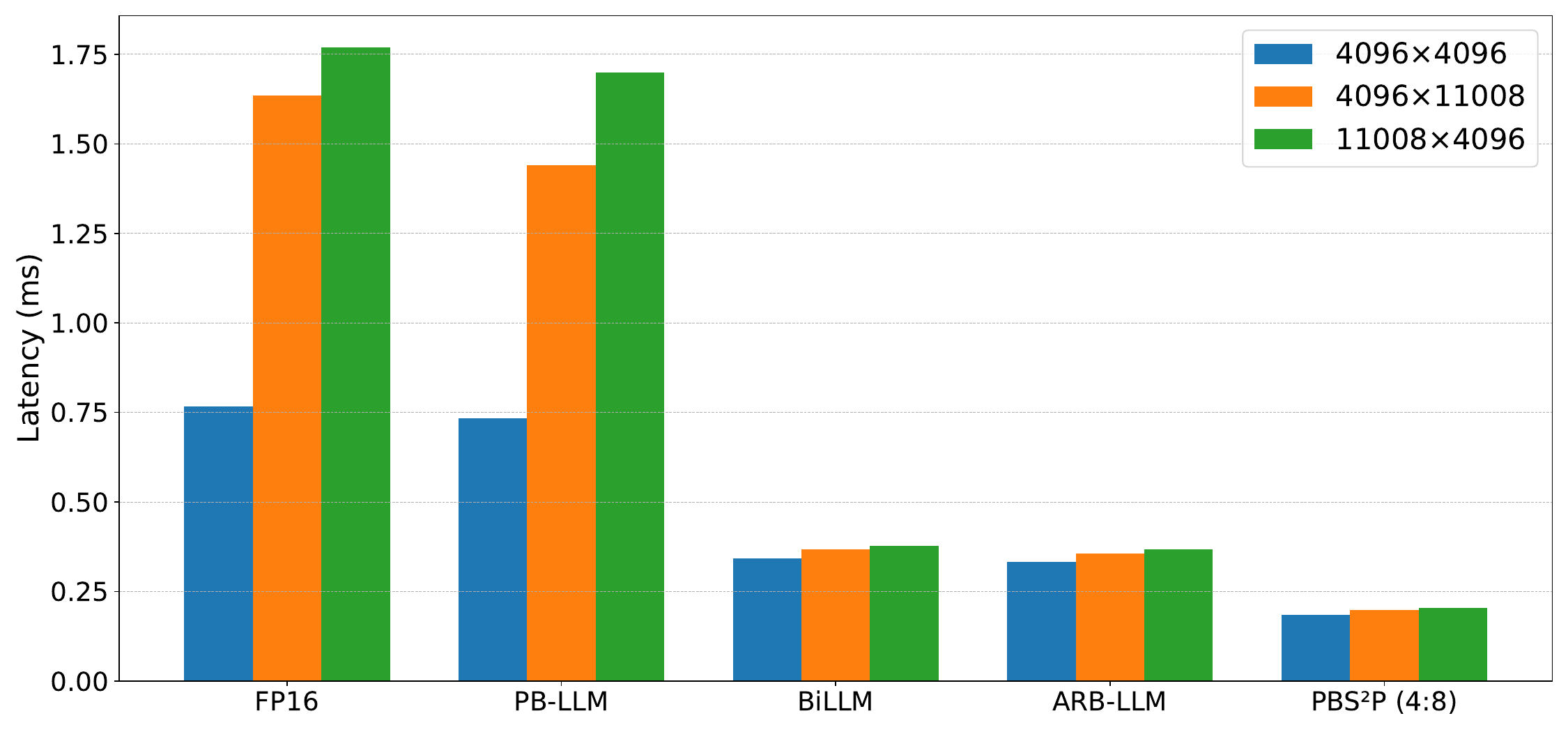}
\vspace{-6mm} 
\caption{Latency comparison (ms) of matrix multiplications across different projection shapes.}
\label{fig:matrix_mul}
\vspace{-6mm}
\end{wrapfigure} 
\vspace{-4mm}
\subsection{Inference Efficiency}
\vspace{-2.5mm}
We evaluate inference efficiency using custom matrix multiplication kernels for binarized and semi-structured sparse weights, built on top of the open-source BitBLAS library\footnote{\ \url{https://github.com/microsoft/BitBLAS}\label{foot:bitblas}}. As shown in the figure~\ref{fig:matrix_mul}, PBS²P (4:8) consistently achieves the lowest latency across typical LLM projection shapes, with activation sequence length fixed at 2048. Compared to FP16 and other 1-bit baselines like BiLLM and ARB-LLM, PBS²P delivers significant speedups, especially on wider matrices such as 4096×11008 and 11008×4096.
\vspace{-8.5mm}
\section{Conclusion}
\vspace{-4.5mm}
In this work, we propose Progressive Binarization with Semi-Structured Pruning (PBS$^2$P) for LLM compression. Central to our approach is SPBO, a stepwise semi-structured pruning strategy with binarization optimization, which prunes a subset of elements at each step while jointly optimizing the binarized parameters. SPBO effectively mitigates the combined error from pruning and binarization. Additionally, we introduce a Coarse-to-Fine Search (CFS) strategy to enhance pruning element selection, further improving compression efficiency. Extensive experiments demonstrate that PBS$^2$P outperforms SOTA binary PTQ methods, delivering superior accuracy across various LLM families and evaluation metrics. Our method enables efficient LLM deployment on resource-limited devices while preserving performance and unifying compression strategies for extreme model reduction.

\newpage


\bibliography{iclr2026_conference}
\bibliographystyle{iclr2026_conference}

\end{document}

%% file: math_commands.tex
\usepackage{amsmath,amsfonts,bm}

\def\eqref#1{equation~\ref{#1}}

\def\1{\bm{1}}

\DeclareMathAlphabet{\mathsfit}{\encodingdefault}{\sfdefault}{m}{sl}
\SetMathAlphabet{\mathsfit}{bold}{\encodingdefault}{\sfdefault}{bx}{n}













%% file: iclr2026_conference.bbl
\begin{thebibliography}{49}
\providecommand{\natexlab}[1]{#1}
\providecommand{\url}[1]{\texttt{#1}}
\expandafter\ifx\csname urlstyle\endcsname\relax
  \providecommand{\doi}[1]{doi: #1}\else
  \providecommand{\doi}{doi: \begingroup \urlstyle{rm}\Url}\fi

\bibitem[An et~al.(2024)An, Zhao, Yu, Tang, and Wang]{an2024fluctuation}
Yongqi An, Xu~Zhao, Tao Yu, Ming Tang, and Jinqiao Wang.
\newblock Fluctuation-based adaptive structured pruning for large language models.
\newblock In \emph{AAAI}, 2024.

\bibitem[Ashkboos et~al.(2024)Ashkboos, Croci, Nascimento, Hoefler, and Hensman]{ashkboos2024slicegpt}
Saleh Ashkboos, Maximilian~L Croci, Marcelo Gennari~do Nascimento, Torsten Hoefler, and James Hensman.
\newblock Slicegpt: Compress large language models by deleting rows and columns.
\newblock In \emph{ICLR}, 2024.

\bibitem[Blumensath \& Davies(2008)Blumensath and Davies]{blumensath2008iterative}
Thomas Blumensath and Mike~E Davies.
\newblock Iterative thresholding for sparse approximations.
\newblock \emph{Journal of Fourier analysis and Applications}, 14:\penalty0 629--654, 2008.

\bibitem[Chakrabarty et~al.(2021)Chakrabarty, Ghosh, Poliak, and Muresan]{chakrabarty2021figurative}
Tuhin Chakrabarty, Debanjan Ghosh, Adam Poliak, and Smaranda Muresan.
\newblock Figurative language in recognizing textual entailment.
\newblock \emph{arXiv preprint arXiv:2106.01195}, 2021.

\bibitem[Chee et~al.(2024)Chee, Cai, Kuleshov, and De~Sa]{chee2024quip}
Jerry Chee, Yaohui Cai, Volodymyr Kuleshov, and Christopher~M De~Sa.
\newblock Quip: 2-bit quantization of large language models with guarantees.
\newblock In \emph{NeurIPS}, 2024.

\bibitem[Chen et~al.(2024)Chen, Lv, Ding, Qin, Zhou, Ding, Liu, Zhang, Guo, Liu, et~al.]{chen2024db}
Hong Chen, Chengtao Lv, Liang Ding, Haotong Qin, Xiabin Zhou, Yifu Ding, Xuebo Liu, Min Zhang, Jinyang Guo, Xianglong Liu, et~al.
\newblock Db-llm: Accurate dual-binarization for efficient llms.
\newblock \emph{arXiv preprint arXiv:2402.11960}, 2024.

\bibitem[Clark et~al.(2019)Clark, Lee, Chang, Kwiatkowski, Collins, and Toutanova]{clark2019boolq}
Christopher Clark, Kenton Lee, Ming-Wei Chang, Tom Kwiatkowski, Michael Collins, and Kristina Toutanova.
\newblock Boolq: Exploring the surprising difficulty of natural yes/no questions.
\newblock In \emph{NAACL}, 2019.

\bibitem[Clark et~al.(2018)Clark, Cowhey, Etzioni, Khot, Sabharwal, Schoenick, and Tafjord]{clark2018think}
Peter Clark, Isaac Cowhey, Oren Etzioni, Tushar Khot, Ashish Sabharwal, Carissa Schoenick, and Oyvind Tafjord.
\newblock Think you have solved question answering? try arc, the ai2 reasoning challenge.
\newblock \emph{arXiv preprint arXiv:1803.05457}, 2018.

\bibitem[Dong et~al.(2024)Dong, Li, Tang, Liu, Pan, Wang, and Chu]{dong2024pruner}
Peijie Dong, Lujun Li, Zhenheng Tang, Xiang Liu, Xinglin Pan, Qiang Wang, and Xiaowen Chu.
\newblock Pruner-zero: Evolving symbolic pruning metric from scratch for large language models.
\newblock In \emph{ICML}, 2024.

\bibitem[Dong et~al.(2025)Dong, Li, Zhong, Du, Fan, Chen, Tang, Wang, Xue, Guo, et~al.]{dong2024stbllm}
Peijie Dong, Lujun Li, Yuedong Zhong, Dayou Du, Ruibo Fan, Yuhan Chen, Zhenheng Tang, Qiang Wang, Wei Xue, Yike Guo, et~al.
\newblock Stbllm: Breaking the 1-bit barrier with structured binary llms.
\newblock In \emph{ICLR}, 2025.

\bibitem[Du et~al.(2024)Du, Zhang, Cao, Guo, Cao, Chu, and Xu]{du2024bitdistiller}
Dayou Du, Yijia Zhang, Shijie Cao, Jiaqi Guo, Ting Cao, Xiaowen Chu, and Ningyi Xu.
\newblock Bitdistiller: Unleashing the potential of sub-4-bit llms via self-distillation.
\newblock In \emph{ACL}, 2024.

\bibitem[Dubey et~al.(2024)Dubey, Jauhri, Pandey, Kadian, Al-Dahle, Letman, Mathur, Schelten, Yang, Fan, et~al.]{dubey2024llama3}
Abhimanyu Dubey, Abhinav Jauhri, Abhinav Pandey, Abhishek Kadian, Ahmad Al-Dahle, Aiesha Letman, Akhil Mathur, Alan Schelten, Amy Yang, Angela Fan, et~al.
\newblock The llama 3 herd of models.
\newblock \emph{arXiv preprint arXiv:2407.21783}, 2024.

\bibitem[Dumitru et~al.(2024)Dumitru, Clotan, Yadav, Peteleaza, and Surdeanu]{dumitru2024change}
Razvan-Gabriel Dumitru, Paul-Ioan Clotan, Vikas Yadav, Darius Peteleaza, and Mihai Surdeanu.
\newblock Change is the only constant: Dynamic llm slicing based on layer redundancy.
\newblock \emph{arXiv preprint arXiv:2411.03513}, 2024.

\bibitem[Frantar \& Alistarh(2022)Frantar and Alistarh]{frantar2022optimal}
Elias Frantar and Dan Alistarh.
\newblock Optimal brain compression: A framework for accurate post-training quantization and pruning.
\newblock In \emph{NeurIPS}, 2022.

\bibitem[Frantar \& Alistarh(2023)Frantar and Alistarh]{frantar2023sparsegpt}
Elias Frantar and Dan Alistarh.
\newblock Sparsegpt: Massive language models can be accurately pruned in one-shot.
\newblock In \emph{ICML}, 2023.

\bibitem[Frantar et~al.(2023)Frantar, Ashkboos, Hoefler, and Alistarh]{frantar2022gptq}
Elias Frantar, Saleh Ashkboos, Torsten Hoefler, and Dan Alistarh.
\newblock Gptq: Accurate post-training quantization for generative pre-trained transformers.
\newblock In \emph{ICLR}, 2023.

\bibitem[Gu et~al.(2024)Gu, Dong, Wei, and Huang]{gu2023knowledge}
Yuxian Gu, Li~Dong, Furu Wei, and Minlie Huang.
\newblock Knowledge distillation of large language models.
\newblock In \emph{ICLR}, 2024.

\bibitem[Hassibi et~al.(1993)Hassibi, Stork, and Wolff]{hassibi1993optimal}
Babak Hassibi, David~G Stork, and Gregory~J Wolff.
\newblock Optimal brain surgeon and general network pruning.
\newblock In \emph{IEEE international conference on neural networks}, pp.\  293--299. IEEE, 1993.

\bibitem[Huang et~al.(2024)Huang, Liu, Qin, Li, Zhang, Liu, Magno, and Qi]{huang2024billm}
Wei Huang, Yangdong Liu, Haotong Qin, Ying Li, Shiming Zhang, Xianglong Liu, Michele Magno, and Xiaojuan Qi.
\newblock Billm: Pushing the limit of post-training quantization for llms.
\newblock In \emph{ICML}, 2024.

\bibitem[Jo et~al.(2024)Jo, Kim, Kim, and Kim]{jo2024mixture}
Dongwon Jo, Taesu Kim, Yulhwa Kim, and Jae-Joon Kim.
\newblock Mixture of scales: Memory-efficient token-adaptive binarization for large language models.
\newblock In \emph{NeurIPS}, 2024.

\bibitem[LeCun et~al.(1989)LeCun, Denker, and Solla]{lecun1989optimal}
Yann LeCun, John Denker, and Sara Solla.
\newblock Optimal brain damage.
\newblock In \emph{NeurIPS}, 1989.

\bibitem[Li et~al.(2021)Li, Gong, Tan, Yang, Hu, Zhang, Yu, Wang, and Gu]{li2021brecq}
Yuhang Li, Ruihao Gong, Xu~Tan, Yang Yang, Peng Hu, Qi~Zhang, Fengwei Yu, Wei Wang, and Shi Gu.
\newblock Brecq: Pushing the limit of post-training quantization by block reconstruction.
\newblock In \emph{ICLR}, 2021.

\bibitem[Li et~al.(2025)Li, Yan, Zhang, Qin, Xie, Tian, Kong, Zhang, Yang, et~al.]{li2024arb}
Zhiteng Li, Xianglong Yan, Tianao Zhang, Haotong Qin, Dong Xie, Jiang Tian, Linghe Kong, Yulun Zhang, Xiaokang Yang, et~al.
\newblock Arb-llm: Alternating refined binarizations for large language models.
\newblock In \emph{ICLR}, 2025.

\bibitem[Lin et~al.(2024)Lin, Tang, Tang, Yang, Chen, Wang, Xiao, Dang, Gan, and Han]{lin2024awq}
Ji~Lin, Jiaming Tang, Haotian Tang, Shang Yang, Wei-Ming Chen, Wei-Chen Wang, Guangxuan Xiao, Xingyu Dang, Chuang Gan, and Song Han.
\newblock Awq: Activation-aware weight quantization for on-device llm compression and acceleration.
\newblock In \emph{MLSys}, 2024.

\bibitem[Liu et~al.(2018)Liu, Wu, Luo, Yang, Liu, and Cheng]{liu2018bi}
Zechun Liu, Baoyuan Wu, Wenhan Luo, Xin Yang, Wei Liu, and Kwang-Ting Cheng.
\newblock Bi-real net: Enhancing the performance of 1-bit cnns with improved representational capability and advanced training algorithm.
\newblock In \emph{ECCV}, 2018.

\bibitem[Liu et~al.(2024)Liu, Oguz, Zhao, Chang, Stock, Mehdad, Shi, Krishnamoorthi, and Chandra]{liu2023llm}
Zechun Liu, Barlas Oguz, Changsheng Zhao, Ernie Chang, Pierre Stock, Yashar Mehdad, Yangyang Shi, Raghuraman Krishnamoorthi, and Vikas Chandra.
\newblock Llm-qat: Data-free quantization aware training for large language models.
\newblock In \emph{ACL}, 2024.

\bibitem[Ma et~al.(2023)Ma, Fang, and Wang]{ma2023llm}
Xinyin Ma, Gongfan Fang, and Xinchao Wang.
\newblock Llm-pruner: On the structural pruning of large language models.
\newblock In \emph{NeurIPS}, 2023.

\bibitem[Marcus et~al.(1994)Marcus, Kim, Marcinkiewicz, MacIntyre, Bies, Ferguson, Katz, and Schasberger]{marcus1994penn}
Mitch Marcus, Grace Kim, Mary~Ann Marcinkiewicz, Robert MacIntyre, Ann Bies, Mark Ferguson, Karen Katz, and Britta Schasberger.
\newblock The penn treebank: Annotating predicate argument structure.
\newblock In \emph{HLT}, 1994.

\bibitem[Merity et~al.(2017)Merity, Xiong, Bradbury, and Socher]{merity2016pointer}
Stephen Merity, Caiming Xiong, James Bradbury, and Richard Socher.
\newblock Pointer sentinel mixture models.
\newblock In \emph{ICLR}, 2017.

\bibitem[Mihaylov et~al.(2018)Mihaylov, Clark, Khot, and Sabharwal]{mihaylov2018can}
Todor Mihaylov, Peter Clark, Tushar Khot, and Ashish Sabharwal.
\newblock Can a suit of armor conduct electricity? a new dataset for open book question answering.
\newblock In \emph{EMNLP}, 2018.

\bibitem[Nvidia(2020)]{nvidia2020}
Nvidia.
\newblock Nvidia a100 tensor core gpu architecture, 2020.

\bibitem[Paszke et~al.(2019{\natexlab{a}})Paszke, Gross, Massa, Lerer, Bradbury, Chanan, Killeen, Lin, Gimelshein, Antiga, et~al.]{paszke1912imperative}
A~Paszke, S~Gross, F~Massa, A~Lerer, JP~Bradbury, G~Chanan, T~Killeen, Z~Lin, N~Gimelshein, L~Antiga, et~al.
\newblock An imperative style, high-performance deep learning library.
\newblock In \emph{NeurIPS}, 2019{\natexlab{a}}.

\bibitem[Paszke et~al.(2019{\natexlab{b}})Paszke, Gross, Massa, Lerer, Bradbury, Chanan, Killeen, Lin, Gimelshein, Antiga, et~al.]{paszke2019pytorch}
Adam Paszke, Sam Gross, Francisco Massa, Adam Lerer, James Bradbury, Gregory Chanan, Trevor Killeen, Zeming Lin, Natalia Gimelshein, Luca Antiga, et~al.
\newblock Pytorch: An imperative style, high-performance deep learning library.
\newblock In \emph{NeurIPS}, 2019{\natexlab{b}}.

\bibitem[Raffel et~al.(2020)Raffel, Shazeer, Roberts, Lee, Narang, Matena, Zhou, Li, and Liu]{raffel2020exploring}
Colin Raffel, Noam Shazeer, Adam Roberts, Katherine Lee, Sharan Narang, Michael Matena, Yanqi Zhou, Wei Li, and Peter~J Liu.
\newblock Exploring the limits of transfer learning with a unified text-to-text transformer.
\newblock \emph{JMLR}, 2020.

\bibitem[Rastegari et~al.(2016)Rastegari, Ordonez, Redmon, and Farhadi]{rastegari2016xnor}
Mohammad Rastegari, Vicente Ordonez, Joseph Redmon, and Ali Farhadi.
\newblock Xnor-net: Imagenet classification using binary convolutional neural networks.
\newblock In \emph{ECCV}, 2016.

\bibitem[Sakaguchi et~al.(2020)Sakaguchi, Le~Bras, Bhagavatula, and Choi]{sakaguchi2019adversarial}
Keisuke Sakaguchi, Ronan Le~Bras, Chandra Bhagavatula, and Yejin Choi.
\newblock Winogrande: An adversarial winograd schema challenge at scale.
\newblock In \emph{AAAI}, 2020.

\bibitem[Sun et~al.(2024)Sun, Liu, Bair, and Kolter]{sun2023simple}
Mingjie Sun, Zhuang Liu, Anna Bair, and J~Zico Kolter.
\newblock A simple and effective pruning approach for large language models.
\newblock In \emph{ICLR}, 2024.

\bibitem[Touvron et~al.(2023{\natexlab{a}})Touvron, Lavril, Izacard, Martinet, Lachaux, Lacroix, Rozi{\`e}re, Goyal, Hambro, Azhar, et~al.]{touvron2023llama1}
Hugo Touvron, Thibaut Lavril, Gautier Izacard, Xavier Martinet, Marie-Anne Lachaux, Timoth{\'e}e Lacroix, Baptiste Rozi{\`e}re, Naman Goyal, Eric Hambro, Faisal Azhar, et~al.
\newblock Llama: Open and efficient foundation language models.
\newblock \emph{arXiv preprint arXiv:2302.13971}, 2023{\natexlab{a}}.

\bibitem[Touvron et~al.(2023{\natexlab{b}})Touvron, Martin, Stone, Albert, Almahairi, Babaei, Bashlykov, Batra, Bhargava, Bhosale, et~al.]{touvron2023llama2}
Hugo Touvron, Louis Martin, Kevin Stone, Peter Albert, Amjad Almahairi, Yasmine Babaei, Nikolay Bashlykov, Soumya Batra, Prajjwal Bhargava, Shruti Bhosale, et~al.
\newblock Llama 2: Open foundation and fine-tuned chat models.
\newblock \emph{arXiv preprint arXiv:2307.09288}, 2023{\natexlab{b}}.

\bibitem[Vaswani(2017)]{vaswani2017attention}
A~Vaswani.
\newblock Attention is all you need.
\newblock In \emph{NeurIPS}, 2017.

\bibitem[Wang et~al.(2023)Wang, Ma, Dong, Huang, Wang, Ma, Yang, Wang, Wu, and Wei]{wang2023bitnet}
Hongyu Wang, Shuming Ma, Li~Dong, Shaohan Huang, Huaijie Wang, Lingxiao Ma, Fan Yang, Ruiping Wang, Yi~Wu, and Furu Wei.
\newblock Bitnet: Scaling 1-bit transformers for large language models.
\newblock \emph{arXiv preprint arXiv:2310.11453}, 2023.

\bibitem[Xia et~al.(2024)Xia, Gao, Zeng, and Chen]{xia2023sheared}
Mengzhou Xia, Tianyu Gao, Zhiyuan Zeng, and Danqi Chen.
\newblock Sheared llama: Accelerating language model pre-training via structured pruning.
\newblock In \emph{ICLR}, 2024.

\bibitem[Xu et~al.(2024)Xu, Han, Yang, Wang, Zhu, Liu, Liu, and Che]{xu2024onebit}
Yuzhuang Xu, Xu~Han, Zonghan Yang, Shuo Wang, Qingfu Zhu, Zhiyuan Liu, Weidong Liu, and Wanxiang Che.
\newblock Onebit: Towards extremely low-bit large language models.
\newblock In \emph{NeurIPS}, 2024.

\bibitem[Yao et~al.(2022)Yao, Yazdani~Aminabadi, Zhang, Wu, Li, and He]{yao2022zeroquant}
Zhewei Yao, Reza Yazdani~Aminabadi, Minjia Zhang, Xiaoxia Wu, Conglong Li, and Yuxiong He.
\newblock Zeroquant: Efficient and affordable post-training quantization for large-scale transformers.
\newblock In \emph{NeurIPS}, 2022.

\bibitem[Yuan et~al.(2023)Yuan, Shang, Song, Wu, Yan, and Sun]{yuan2023asvd}
Zhihang Yuan, Yuzhang Shang, Yue Song, Qiang Wu, Yan Yan, and Guangyu Sun.
\newblock Asvd: Activation-aware singular value decomposition for compressing large language models.
\newblock \emph{arXiv preprint arXiv:2312.05821}, 2023.

\bibitem[Zellers et~al.(2019)Zellers, Holtzman, Bisk, Farhadi, and Choi]{zellers2019hellaswag}
Rowan Zellers, Ari Holtzman, Yonatan Bisk, Ali Farhadi, and Yejin Choi.
\newblock Hellaswag: Can a machine really finish your sentence?
\newblock In \emph{ACL}, 2019.

\bibitem[Zhang et~al.(2024)Zhang, Chen, Shen, Yang, Ou, Yu, and Zhuang]{zhang2023loraprune}
Mingyang Zhang, Hao Chen, Chunhua Shen, Zhen Yang, Linlin Ou, Xinyi Yu, and Bohan Zhuang.
\newblock Loraprune: Pruning meets low-rank parameter-efficient fine-tuning.
\newblock In \emph{ACL}, 2024.

\bibitem[Zhang et~al.(2022)Zhang, Roller, Goyal, Artetxe, Chen, Chen, Dewan, Diab, Li, Lin, et~al.]{zhang2022opt}
Susan Zhang, Stephen Roller, Naman Goyal, Mikel Artetxe, Moya Chen, Shuohui Chen, Christopher Dewan, Mona Diab, Xian Li, Xi~Victoria Lin, et~al.
\newblock Opt: Open pre-trained transformer language models.
\newblock \emph{arXiv preprint arXiv:2205.01068}, 2022.

\bibitem[Zhong et~al.(2024)Zhong, Ding, Shen, Liu, Du, and Tao]{zhong2024revisiting}
Qihuang Zhong, Liang Ding, Li~Shen, Juhua Liu, Bo~Du, and Dacheng Tao.
\newblock Revisiting knowledge distillation for autoregressive language models.
\newblock In \emph{ACL}, 2024.

\end{thebibliography}
